\pdfoutput=1
\documentclass[11pt,a4paper]{article}
\usepackage[hyperref]{acl}
\usepackage{times}
\usepackage{latexsym}
\usepackage{danudefs}
\usepackage{algorithmic}
\usepackage{algorithm}
\usepackage{color}
\usepackage{booktabs}
\usepackage{xspace}
\usepackage{url}
\usepackage{microtype}
\usepackage{wasysym}
\usepackage{csquotes}
\usepackage{arydshln}
\usepackage{multirow}
\usepackage{subcaption}
\usepackage{graphics}

\usepackage[english]{babel}
\usepackage{hyperref}
\addto\captionsenglish{%
}
\addto\extrasenglish{%
}

\usepackage{xr-hyper}
\makeatletter
\newcommand*{\addFileDependency}[1]{
  \typeout{(#1)}
  \@addtofilelist{#1}
  \IfFileExists{#1}{}{\typeout{No file #1.}}
}
\makeatother
\newcommand*{\myexternaldocument}[1]{%
    \externaldocument{#1}%
    \addFileDependency{#1.tex}%
    \addFileDependency{#1.aux}%
}
\myexternaldocument{Appendices}

\usepackage{todonotes}

\makeatletter
\if@todonotes@disabled

\else

\fi
\makeatother

\title{\emph{Learning to Borrow} -- Relation Representation for \\ Without-Mention Entity-Pairs for Knowledge Graph Completion}

\author{ Huda Hakami$^{1}$, Mona Hakami$^{2}$, Angrosh Mandya$^{3}$ \and Danushka Bollegala$^{3,4}$ \\
$^1$Department of Computer Science, Taif University, Saudi Arabia \\
$^2$King Saud University, Saudi Arabia \\
$^3$University of Liverpool, UK \\
$^4$Amazon \\\texttt{hahakami@tu.edu.sa}, \texttt{mohakami@ksu.edu.sa}\\
\texttt{\{angrosh,danushka\}@liverpool.ac.uk}}

\date{}
\graphicspath{{./Figures/}}
\begin{document}
\maketitle

\begin{abstract}

Prior work on integrating text corpora with knowledge graphs (KGs) to improve Knowledge Graph Embedding (KGE) have obtained good performance for entities that co-occur in sentences in text corpora.
Such sentences (textual mentions of entity-pairs) are represented as Lexicalised Dependency Paths (LDPs) between two entities. 
However, it is not possible to represent relations between entities that do not co-occur in a single sentence using LDPs.
In this paper, we propose and evaluate several methods to address this problem, where we \emph{borrow} LDPs from the entity pairs that co-occur in sentences in the corpus (i.e. \emph{with mention} entity pairs) to represent entity pairs that do \emph{not} co-occur in any sentence in the corpus (i.e. \emph{without mention} entity pairs).
We propose a supervised borrowing method, \emph{SuperBorrow}, that learns to score the suitability of an LDP to represent a without-mention entity pair using pre-trained entity embeddings and contextualised LDP representations.
Experimental results show that SuperBorrow improves the link prediction performance of multiple widely-used prior KGE methods such as TransE, DistMult, ComplEx and RotatE. 
\end{abstract}

\section{Introduction}
Knowledge Graphs (KGs) are a structured form of information that underline the relationships between real-world entities~\citep{ehrlinger2016towards,kroetsch2016special,paulheim2017knowledge}. 
A KG is represented using a set of relational tuples of the form $(h, r, t)$, where $r$ represents the relation between the head entity $h$ and the tail entity $t$. 
For example, the relational tuple (\emph{Joe Biden}, \emph{president-of}, \emph{US}) indicates that the \emph{president-of} relation holds between \emph{Joe Biden} and \emph{US}. 
There exists a large number of publicly available and widely used KGs, such as Freebase~\citep{bollacker2008freebase}, DBpedia~\citep{auer2007dbpedia}, and YAGO ontology~\citep{suchanek2007yago}.
KGs have been effectively applied in various NLP tasks such as, relation extraction~\citep{riedel2013relation,weston2013connecting}, question answering~\citep{das2017question,sydorova2019interpretable}, and dialogue systems~\citep{xu2020knowledge}.
However, most KGs suffer from data sparseness as many relations between entities are not explicitly represented~\citep{min2013distant}.

To overcome the sparsity problem, Knowledge Graph Embedding (KGE) models learn representations (a.k.a. embeddings) for entities and relations in a given KG in a vector space, which can then be used to infer missing links between entities~\citep{bordes2013translating,nickel2015review,wang2017knowledge}. 
Such models are trained to predict relations that are likely to exist between entities (known as link prediction or KG completion) according to some scoring formula. 
Although previously proposed KGE methods have shown good empirical performances for KG completion~\citep{minervini2015efficient}, the KGEs are learnt from the KGs only, which might not represent all the relations that exist between the entities included in the KG.  
To overcome this limitation, prior work has used external text corpora in addition to the KGs~\citep{toutanova2015representing,xu2016knowledge,long2016leveraging,an2018accurate,wang2019model,wang2019knowledge,lu2020utilizing}.  
Compared to structured KGs, unstructured text corpora are abundantly available, up-to-date and have diverse linguistic expressions for extracting relational information.  

The co-occurrences of two entities within sentences (a.k.a textual mentions) in a text corpus has shown its success for text-enhanced KGEs~\citep{komninos2017feature,an2018accurate}. For example, the relational tuple in the Freebase KG (\emph{Joe Biden}, \emph{president-of}, \emph{US}) is mentioned in the following sentence \enquote{Joseph Robinette Biden Jr. is an American politician who is the 46th and current president of the United States.}
This sentence expresses the \emph{president-of} relation between the two entities \emph{Joe Biden} and \emph{US}.
As the entity-pair (\emph{Joe Biden},\emph{US}) appears in a single sentence, we call it a \emph{with-mention} entity-pair. 
However, even in a large text corpus, not every related entity pair co-occurs in a specified window, which are referred to as \emph{without-mention} entity-pairs in previous studies.
For instance, if we consider the widely used FB15K-237 KG~\citep{toutanova2015representing} and the ClueWeb12~\citep{gabrilovich2013facc1} text corpus with FB entity mention annotations,\footnote{200 million sentences in CluWeb12 annotated with FB entity mention annotations.} 33\% of entity-pairs in FB15k-237 do not have textual mentions within the same sentences. 
This sparseness problem limits the generalisation capabilities of using textual mentions for enhancing KGEs.
Specifically,~\citet{toutanova2015representing,komninos2017feature} have shown larger improvements in link prediction for with-mention entity-pairs over without-mention pairs.  

In this paper, we propose a method to augment a given KG with additional textual relations extracted from a corpus and represented as LDPs. 
The augmented KG can then be used to train \emph{any} KGE learning method. 
This is attractive from both scalability and compatibility point of views because our proposal is agnostic to the KGE learning method that is subsequently used for learning KGEs.
Our main contribution in this paper is to improve link prediction for without-mention entity-pairs by borrowing LDPs from with-mention entity-pairs to overcome the sparseness in co-occurrences of the without-mention entity-pairs. 
We show that learning a supervised borrowing method, \emph{SuperBorrow}, that scores suitable LDPs to represent without-mention entity-pairs based on pre-trained entity embeddings and contextualised LDP embeddings boosts the performance of link prediction
using a series of KGE methods, compared to what would have been possible without textual relations.

\section{Related Work}

\noindent\textbf{KGEs from a Multi-relational Graph:} Typically, KG embedding models consist of two main steps: (a) defining a scoring function for a tuple, and (b) learning entity and relation representations. 
Entities are usually represented as vectors, whereas relations can be represented by vectors (e.g. TransE~\citep{bordes2013translating}, DistMult~\citep{yang2014embedding} and ComplEx~\citep{trouillon2016complex}) matrices (e.g. RESCAL~\citep{nickel2011three}), or by 3D tensors (e.g. Neural Tensor Network~\citep{socher2013reasoning}). 

Using some form of a representation, scoring functions are then defined to evaluate the strength of a relation $r$ between $h$ and $t$ entities in a triple. 
TransE is one of the earliest and well-known distance-based KGE method that performs a linear translation and its scoring function is given in \autoref{tbl:scores}.  
Alternatively, a bilinear function is used in several KGE models, such as RESCAL, DistMult and ComplEx, for which scoring functions are defined in  \autoref{tbl:scores}.
KGEs are learnt such that the observed facts (positive triples) are assigned higher scores compared to that of the negative triple (for example generated by perturbing a positive instance by replacing its head or tail entities by an entity randomly selected from the set of entities) by minimising a loss function, such as the logistic loss or the margin loss. 

Conventional KGE models are trained using the facts in the KGs, which are often incomplete. 
Therefore, to overcome the sparsity of structured KGs, we propose to integrate information from a text corpus, thereby augmenting the KG.
The augmented KG is then used as the input to existing KGE methods to learn accurate entity and relation embeddings. In particular, we do not modify the scoring functions nor optimisation objectives for the respective KGE methods, which makes our proposed approach applicable in many existing KGE methods without any modifications.

\begin{table}[t!]
\centering
\scalebox{0.82}{
\begin{tabular}{l l}
\toprule
KGE method & Score function  \\ 
 &  $f(h, R, t)$  \\ \midrule
TransE\small{~\cite{bordes2013translating}} & $\norm{\vec{h} + \vec{r} - \vec{t}}_{\ell_{1/2}}$ \\
DistMult\small{~\cite{yang2014embedding}} & $\langle\vec{h}, \vec{r}, \vec{t}\rangle$  \\
ComplEx\small{~\cite{trouillon2016complex}}  &  $\langle\vec{h}, \vec{r}, \bar{\vec{t}}\rangle$  \\
RotatE\small{~\cite{sun2019rotate}} & $\norm{\vec{h}\circ\vec{r}-\vec{t}}^2$  \\
\bottomrule
\end{tabular}}

\caption{Score functions proposed in KGE methods. Entity embeddings $\vec{h}, \vec{t} \in \R^{d}$ are vectors in all models, except in ComplEx where $\vec{h},\vec{t} \in \mathbb{C}^{d}$. Here, $\ell_{1/2}$ denotes either $\ell_{1}$ or $\ell_{2}$ norm of a vector. In ComplEx, $\bar{\vec{t}}$ is the element-wise complex conjugate.}
\label{tbl:scores}
\end{table}

\noindent\textbf{Text-Enhanced KGEs:} 
Recently, a new line of research that combines textual information with relational graphs has emerged~\citep{lu2020utilizing}. 
Different combination methods have been proposed for this purpose.
\citet{wang2014knowledge} proposed a model to embed both entities and words (using entity names and Wikipedia anchors) into the same low-dimensional vector space to capture relational information from a KG and the co-occurrences from the corpus.
\citet{rosso2019revisiting} control the amount of information shared between the two data sources in the joint embedding space using regularisation. 
This joint model is further enhanced by incorporating entity descriptions from an external corpus, which are jointly learnt with the KG~\citep{zhong2015aligning,xie2016representation,veira2019unsupervised}.
In a different scenario, the text-enhanced knowledge embedding model by~\citet{wang2016text} creates a co-occurrence network of words and entities from an entity-annotated corpus. The authors define point-wise and pair-wise contexts using the co-occurrence frequencies in the network. 
Then, entity and relation embeddings are enhanced using textual point-wise and pair-wise embeddings, respectively.
Similarly,~\citet{rezayi-etal-2021-edge} construct an augmented KG that has nodes from external text.
The original and the augmented graphs are then aligned to suppress the noise and distil relevant information. 
In our work, we focus on adding extra edges to the KG rather than nodes as in~\citet{rezayi-etal-2021-edge} and ~\citet{wang2016text}. 

In addition to contextual information and textual descriptions of individual words/entities, sentences where two entities co-occur  have been used as contextual evidence to learn KGEs~\citep{toutanova2015representing,komninos2017feature,tang2019knowledge}. 
For example,~\citet{toutanova2015representing} extracted LDPs by parsing co-occurring sentences in a text corpus, which are then used as textual relations in the KG. 
This model can be seen as a special case of \emph{universal schema}~\citep{riedel2013relation}, which combines textual and KG relations in the same entity-pair co-occurrence matrix, subsequently decomposed to obtain entity embeddings. 
\citet{komninos2017feature} proposed a novel triple scoring function where textual mentions are used as a source of supporting evidence for a triple.


Our problem setting differs from  prior work on text-enhanced KGEs in two important ways. 
First, we do not modify the underlying structure of the KGE method, which is attractive from both scalability and compatibility of our proposal.
Second, rather than considering only entity-pairs that are occurring within a specified context in the corpus (i.e. with-mention entity-pairs), we propose to borrow LDPs from with-mention entity-pairs to overcome the data sparseness in without-mention entity-pairs that never co-occur within any sentence in the corpus.

\section{Method}

A relational KG $\cD$ consists of a set of entities $\cE$ and a set of relations $\cR$. In $\cD$, knowledge is represented by relational tuples $(h,r,t)\in\cD$, where the head entity $h$ is related to the tail entity $t$ by the KG relation $r$. 
In this work, we assume relations to be asymmetric in general (if $(h, r, t) \in \cD$ then it does not necessarily follow that $(t, r, h) \in \cD)$. 
The goal is to learn representations for entities and relations such that missing tuples can be accurately inferred. 

As KGs $\cD$ are often sparse with many missing edges between entities, the learnt KGEs are affected, which in return impacts the performance of KGEs on downstream tasks such as link prediction. 
To address this sparseness problem, we consider the availability of a text corpus $\cT$ where relational facts are expressed using contexts in which an entity-pair co-occurs.
The textual relations that are extracted from $\cT$ can be \emph{injected} into $\cD$ before applying a KGE method. 

To align $\cD$ with $\cT$, entity linking is applied to resolve ambiguous entity mentions in the text with unique entities in the KG~\citep{gabrilovich2013facc1,shen2014entity}. 
Then, Sentences that mentions two entities are considered as \emph{textual mentions} of relations between entities. 
Assuming that the corpus is annotated using the entities in $\cD$, there are multiple possibilities to obtain relational features of sentences that mention the entities. 
Following previous work~\cite{toutanova2015representing}, we first run a dependency parser~\citep{chen2014fast} on each sentence containing an entity-pair to obtain LDPs. 
Then, if $\cE$ contains the head and tail entities of an LDP $l$ (but the entity-pair might not be connected by KG relations), we insert $l$ into $\cR$ to form a textual triple $(h,l,t)\in\cD$. 
The augmented KG is then used to learn embeddings for $\cE$ and $\cR$ using different KGE methods. 
During KGE processs, we treat both original relations in the KG and the augmented LDPs equally.
In principle, any existing KGE learning method can be applied on the augmented KG as we later see in our experiments.

One obvious limitation of the above-described method is that entity-pairs that never co-occur in any contextual window (e.g. a sentence) will \emph{not} be connected by any LDP during the augmentation process.
This is fine if the two entities are truly unrelated.
However, this is problematic for entities that are related in the KG but their relations were not sufficiently covered in the text corpus because of the coverage issues and small size of the corpus.
As we later see in our evaluations (\autoref{sec:Res}), this is indeed the case for the majority of the without-mention entity-pairs.
To overcome this limitation of our proposal, next we describe a method to \emph{borrow} LDPs from with-mention entity-pairs to without-mention entity pairs.
It is worth noting that we do not connect any two entities by LDPs, but only those that are related in the KG and predicted to be associated with an LDP by the proposed method.  


\subsection{Learn to Borrow LDPs}

Given a without-mention entity pair $(h^*,t^*)$, we propose a supervised borrowing method \textbf{SuperBorrow} to rank LDPs that are extracted for the with mention entity-pairs from a text corpus.
Given pre-trained entity representations $\vec{h}$ and $\vec{t}$, we learn an entity-pair encoder, $f$, parametrised by $\theta$, to create an entity-pair representation, $f(h,t; \theta)$, for $(h,t)$. 
In this work, the encoder $f$ is implemented as a multilayer perceptron with a nonlinear activation, where the input entity-pair to the MLP is encoded as follows:
\begin{align}
\label{eq:inputlayer}
    \vec{x}=[\vec{h}\oplus\vec{t}\oplus(\vec{h}-\vec{t})\oplus(\vec{h}\circ\vec{t})]
\end{align}
Here, $\oplus$ denotes the concatenation of vectors and $\circ$ is the element-wise multiplication between two vectors.  
\eqref{eq:inputlayer} considers the information in the head and tail entity embeddings independently as well as the interactions between their corresponding dimensions.
These features for entity-pairs have been used successfully for representing semantic relations in prior work~\cite{washio2018neural,joshi2018pair2vec,hakami2019context}. 
The final output vector $f(h,t;\theta)$ of the MLP is treated as the representation of the entity-pair $(h,t)$.

As an alternative to representing the relationship between two entities in an entity-pair $(h,t)$ by $f(h,t;\theta)$ using the corresponding entity embeddings, we can use $\cS_{(h,t)}$, the set of LDPs connecting $h$ and $t$ entities~\cite{Bollegala:WWW:2010}.
Because an LDP is a sequence of textual tokens, we can use any sentence encoder to represent an LDP by a vector.
Specifically, in our experiments later we use the pretrained sentence encoder SBERT~\cite{reimers2019sentence} to represent an LDP, $l$, by a vector, $\vec{l}$. 

\begin{table}[]
    \centering
     \scalebox{0.75}{
    \begin{tabular}{l c c c c c}\toprule
         &Relations&Entities&Triples&w-m&w/o-m  \\
         &&&Train/Test&&\\
         \midrule
         FB&237&14,541&272,115/20,466&2,344&18,122\\
         Text&1,100&12,930&404,009/-&-&-\\
         \bottomrule
    \end{tabular}}
    \caption{Statistics of the datasets. w-m and w/o-m denotes the number of test instances respectively  in with-mention and  without-mention entity-pair sets.}
    \label{tab:stat}
\end{table}

We require LDPs that co-occur with an entity-pair $(h,t)$ to be similar to $f(h,t;\theta)$ than LDPs that do not co-occur with $(h,t)$.
Specifically, we use the set of with-mention entity-pairs with their associated LDPs as positive training instances $\cS=\{(h,l,t)\}$.
LDPs that are associated with either $h$ or $t$ alone (not both) are used as negative training instances $\cS'_{(h,t)}$ as given by \eqref{eq:neg}.
\par

\begin{align} 
\label{eq:neg}
\vspace{-5mm}
\cS'_{(h,t)} = \{&(h,l',t)| \exists t':(h,l',t')\in\cD \wedge t' \neq t, \nonumber \\  
             & \exists h':(h',l',t)\in\cD \wedge h' \neq h\}
\end{align}

We learn the parameters of $f(h,t,\theta)$ by minimising the marginal loss over $\cS_{(h,t)}$ and $\cS'_{(h,t)}$ as shown in~\eqref{eq:loss}.
\par

\begin{align}
\label{eq:loss}
\vspace{-5mm}
    \sum_{(h,l,t)\in\cS_{}} \sum_{(h,l',t)\in\cS'_{(h,t)}}  \max\left(0,\gamma - f(h,t;\theta)\T (\vec{l} - \vec{l'})\right)
\end{align}

\begin{table*}[t]
\small

        \scalebox{0.82}{
        \begin{tabular}{l c c c c c  c c c c c c c c c c} \toprule
			 & \multicolumn{5}{c}{overall} & \multicolumn{5}{c}{with-mention}&\multicolumn{5}{c}{without-mention} \\ 
			 \cmidrule(lr){2-6}  \cmidrule(lr){7-11} \cmidrule(lr){12-16} \\
			Model& MRR& MR & H@10&H@3&H@1 & MRR& MR & H@10&H@3&H@1 & MRR& MR & H@10&H@3&H@1\\ \midrule
			\textbf{TransE} (KG only) &0.336&	113	&0.523&	0.368&	0.243&	0.314&	135&	0.508&	0.349&	0.218&	0.333&	111&	0.519&	0.364&	0.241\\ 
			KG+ExtractedLDPs & 0.314	&126&	0.495&	0.343&	0.224&	\textbf{0.433}&	\textbf{43}&	\textbf{0.659}&	0.489&	\textbf{0.319}&	0.293&	138&	0.468&	0.318&	0.206\\
			\hdashline
			LinkAll &0.344&	105&	0.531&	0.380&	0.249&	0.430&	44&	0.653&	\textbf{0.493}&	0.316&	0.328&	113&	0.510&	0.360&	0.235 \\
			Co-occurrence&0.502&	47&	0.695&	0.553&	0.402&	0.412&	48&	0.639&	0.464&	0.297&	0.506&	47&	0.695&	0.557&	0.408 \\
			NeighbBorrow&0.491&	49&	0.682&	0.541&	0.392&	0.422&	46&	0.646&	0.475&	0.308&	0.493&	50&	0.68&	0.542&	0.395\\
			SuperBorrow&\textbf{0.751}&	\textbf{15}&	\textbf{0.868}&	\textbf{0.799}&	\textbf{0.681}&	0.394&	49&	0.629&	0.445&	0.277&	\textbf{0.787}&	\textbf{11}&	\textbf{0.888}&	\textbf{0.835}&	\textbf{0.723}\\
			\midrule
			\\
			\textbf{DistMult} (KG only)& 0.302&	133&	0.489&	0.333&	0.209&	0.257&	149&	0.436&	0.289&	0.165&	0.302&	131&	0.489&	0.333&	0.209\\
			KG+ExtractedLDPs&0.325&	113&	0.512&	0.357&	0.232&	0.427&	35&	0.656&	0.483&	0.311&	0.306&	125&	0.488&	0.335&	0.216\\
			\hdashline
			LinkAll & 0.329&	108&	0.521&	0.363&	0.233&	\textbf{0.437}&	\textbf{33}&	\textbf{0.670}&	\textbf{0.496}&	\textbf{0.315}&	0.309&	118&	0.497&	0.339&	0.215 \\
			Co-occurrence&0.365	&74	&0.574	&0.404	&0.261	&0.428	&\textbf{33}	&0.664	&0.479	&0.310	&0.351	&81	&0.558	&0.388	&0.248 \\
			NeighbBorrow&0.415&	54&	0.639&	0.465&	0.302&	0.412&	35&	0.645&	0.463&	0.297&	0.408&	57&	0.631&	0.458&	0.295\\
			SuperBorrow& \textbf{0.482}&	\textbf{53}&	\textbf{0.681}&	\textbf{0.535}&	\textbf{0.377}&	0.415&	35&	0.655&	0.475&	0.291&	\textbf{0.482}&	\textbf{56}&	\textbf{0.678}&	\textbf{0.534}&	\textbf{0.379}\\
			\midrule
			\\
			\textbf{ComplEx} (KG only)&0.312&	125&	0.493&	0.342&	0.222&	0.275&	142&	0.459&	0.299&	0.185&	0.312&	124&	0.493&	0.342&	0.222\\
			KG+ExtractedLDPs&0.321&	107&	0.505&	0.349&	0.229&	0.407&	36&	0.637&	0.458&	0.291&	0.304&	117&	0.482&	0.329&	0.216\\
			\hdashline
			LinkAll &0.328&	107&	0.519&	0.361&	0.232&	0.432&	34&	0.665&	0.493&	0.311&	0.309&	118&	0.496&	0.338&	0.216 \\
			Co-occurrence&0.358&	76&	0.570&	0.399&	0.252&	\textbf{0.436}&	\textbf{33}&	\textbf{0.679}&	\textbf{0.499}&	\textbf{0.319}&	0.342&	83&	0.552&	0.380&	0.238 \\
			NeighbBorrow&0.428&	47&	0.650&	0.479&	0.315&	0.418&	35&	0.646&	0.478&	0.298&	0.422&	50&	0.643&	0.472&	0.309\\
			SuperBorrow&\textbf{0.489}&	\textbf{42}&	\textbf{0.687}&	\textbf{0.540}&	\textbf{0.385}&	0.416&	38&	0.653&	0.481&	0.291&	\textbf{0.491}&	\textbf{43}&	\textbf{0.686}&	\textbf{0.541}&	\textbf{0.388}\\
		    \midrule
		    \\
			\textbf{RotatE} (KG only) &0.358&	94&	0.560&	0.395&	0.259&	0.331&	120&	0.527&	0.365&	0.236&	0.354&	92&	0.557&	0.391&	0.254\\
			KG+ExtractedLDPs&0.359&	94&	0.551&	0.396&	0.264&	\textbf{0.448}&	\textbf{44}&	0.672&	\textbf{0.509}&	\textbf{0.333}&	0.341&	101&	0.528&	0.374&	0.247\\
			\hdashline
			LinkAll &0.363&	91&	0.558&	0.400&	0.266&	0.442&	\textbf{44}&	0.671&	0.503&	0.321&	0.346&	98&	0.536&	0.378&	0.251 \\
			Co-occurrence&0.435&	54&	0.639&	0.484&	0.329&	0.441&	46&	0.663&	0.499&	0.327&	0.426&	56&	0.629&	0.473&	0.321 \\
			NeighbBorrow&0.463&	43&	0.672&	0.514&	0.357&	0.443&	45&	\textbf{0.675}&	0.497&	0.326&	0.457&	44&	0.664&	0.508&	0.351\\
			SuperBorrow&\textbf{0.682}&	\textbf{19}
			&\textbf{0.836}&	\textbf{0.739}&	\textbf{0.594}&	0.412&	51&	0.652&	0.473&	0.290&	\textbf{0.706}&	\textbf{16}&	\textbf{0.851}&	\textbf{0.764}&	\textbf{0.621}\\

			\bottomrule
	\end{tabular}}
 
    \caption{Results of link prediction on FB15K237. Higher is better for all metrics except for the mean rank (MR) for which lower values indicate better performance. The best result for each metric and each KGE method is shown in bold.}
    \label{tab:LinkPredict}
\end{table*}

Here, $\gamma (\geq 0)$ is the margin and is set to 1 in our experiments. 
To determine which LDPs to be borrowed for a particular without-mention entity pair, $(h^*,t^*)$, we first compute its representation, $f(h^*,t^*; \theta)$ using the $\theta$ found by minimising \eqref{eq:loss} above.
We then score each LDP, $l$, using the sentence encoder model, by the inner-product, $f(h^*,t^*; \theta)\T\vec{l}$.
We then select the top-$k$ LDPs with the highest inner-products with $f(h^*,t^*; \theta)$ to augment the KG.
The number of borrowed LDPs ($k$) is a hyperparameter that is tuned using the validation triples selected from the KG.

\section{Experimental Setup}
\label{sec:exp}


\subsection{Dataset and Training Details}
\label{sec:exp-settings}

\noindent\textbf{Datasets:} We use FB15k237 as the KG and  ClueWeb12\footnote{\url{https://lemurproject.org/clueweb12/}} as the corpus for extracting LDPs for the entity-pairs in the FB157k237 KG. 
Specifically, we use the textual triples consisting of LDPs that are extracted and made available\footnote{\url{https://www.microsoft.com/en-us/download/details.aspx?id=52312}} by~\citet{toutanova2015representing}.
The number of extracted unique LDPs and textual triples in this dataset are respectively $2,740,176$ and $3,978,014$.   
To make the training of KGE methods computationally efficient, we filter out LDPs that occur in less than $100$ distinct entity-pairs in the corpus. 
The FB15k237 test set is split into with-mention (i.e. entity-pairs that co-occur in some LDP) and  without-mention (i.e entity-pairs that do not co-occur in any LDP) sets as shown in~\autoref{tab:stat}. 
According to \autoref{tab:stat}, there are $88.14\%$ without-mention entity-pairs in the test set. 
Note that even if we consider the complete set of LDPs from the ClueWeb12, the portion of without-mention test entity-pairs in FB15k237 is only $73.23\%$. 
This shows the significance of the problem of representing without-mention entity-pairs, which is the focus of this paper.

\noindent\textbf{Training SuperBorrow:} 
We used the with-mention entity-pairs in train split of FB15K237 to train SuperBorrow. 
The number of entity-pairs in the training set is 311,906, and on average we have $1.32$ LDPs per entity-pair.
On average, we generate 100 negative triples for each with-mention pair.
We hold-out $10\%$ of the training entity-pairs for validation purposes (we obtain $280716$ and $31190$ entity-pairs for training and validation, respectively).
To represent each entity, we use the publicly available 100-dimensional pre-trained RelWalk embeddings,\footnote{\url{https://github.com/LivNLP/Relational-Walk-for-Knowledge-Graphs}}
which are publicly available for the entities and relations in FB15k237.

According to~\eqref{eq:inputlayer}, the input layer of the trained MLP has 400 features. 
The hyperparameters including the number of hidden layers $\{2,3\}$, $\ell_{2}$, regularisation coefficient $\{0,0.01,0.001\}$, the learning rate $\{0.01,0.1\}$ and the non-linear activation $\{\textrm{tanh},\textrm{relu}, \textrm{sigmoid}\}$ are tuned using the above-mentioned validation set. 
The MLP consists of two 768-dimensional layers, and the last layer represents the entity-pair to be mapped to the LDP embedding space that has 768 dimensions encoded using the SBERT \texttt{paraphrase-distilroberta-base}\footnote{\url{https://huggingface.co/sentence-transformers/paraphrase-distilroberta-base-v2}} model, which has reported SoTA performance on various knowledge-intensive tasks~\citep{warstadt2020learning}. 
SuperBorrow is trained for 50 iterations using mini-batch Stochastic Gradient Descent with momentum and a batch size of 128. 
The source for SuperBorrow is publicly available.\footnote{\url{https://github.com/Huda-Hakami/Learning-to-Borrow-for-KGs}} 

\noindent\textbf{Evaluation Protocol:} 
After augmenting FB15K237 with the borrowed $k$ LDPs for each without-mention entity-pair, we train a KGE method to obtain embeddings for the entities in $\cE$, relations in $\cR$ and textual relations.
The hyperparameter $k$ is tuned on the validation set of FB15K237 for each KGE method from $\{1,3,10,15,20,25,30\}$.

We use Link Prediction, which has been popularly used as an evaluation task to compare the KGEs we obtain from a KGE learning method before and after augmenting the KG with the LDPs borrowed using SuperBorrow and other baselines~\citep{wang2021survey}. 
Link prediction is the task of predicting the missing head (i.e. $(? , r, t)$) or tail (i.e. $(h,r,?)$) entity in a given triple by ranking the entities in the KG according to the scoring function of the KGE method.
Following prior work, the performance is evaluated using Mean Reciprocal Rank (MRR), Mean Rank (MR) and Hits at Rank k (H@k) under the \emph{filtered} setting, which removes all triples appeared in training, validating or testing sets from candidate triples before obtaining the rank for the ground truth triple.
We consider all entities that appear in the corresponding argument of the relation to be predicted to further filter out incorrect candidates, which is known as type-constraint setting~\citep{chang2014typed,toutanova2015observed}.

We also evaluate the learnt KGEs using a relation prediction task, which predicts the relation between two given entities (i.e., $(h,?,t)$) from the set of relations in the KG. 
This task assumes that we are given entity-pairs with candidate relations. 
The performance is measured using the same evaluation metrics as used in the link prediction task under the filtered setting. 
We use the publicly available OpenKE tool to conduct experiments with different KGE methods~\citep{han2018openke}.\footnote{\url{https://github.com/thunlp/OpenKE}}

\subsection{Baselines}
We compare the proposed LDP borrowing method against multiple baselines as follows.

\paragraph{LinkAll:} 
In this baseline we connect the two entities in each without-mention entity-pair with a unique link, instead of reusing LDPs, and augment the KG with the without-mention entity-pairs.
This baseline enables us to simply incorporate all without-mention entity-pairs into the KG without requiring to borrow any LDPs.
It will demonstrate the importance, if any, of sharing LDPs between with- vs. without-mention entity-pairs as opposed to simply connecting all without-mention entity-pairs with distinct relations.


\paragraph{Co-occurrence:} 
This baseline connects all entity-pairs that co-occurs in any sentence in the corpus ($\cT$) with a generic relation (i.e. co-occurrence relation) in the augmented KG and does not distinguish between different textual relations.
This baseline is designed to highlight the importance of the context of entity-pair co-occurrences in the corpus beyond simply treating all co-occurrences equally during the augmenting process.

\paragraph{NeighbBorrow:} Given a without-mention entity-pair $(h^*,t^*)$, we can borrow the LDPs from the first nearest neighbouring (1NN) with-mention entity-pair $(h,t)$. 
The similarity between entity-pairs can be computed using \eqref{eq:entitypairs} in an unsupervised manner using pretrained entity embeddings such as RelWalk embeddings~\citep{bollegala-etal-2021-relwalk}.
\begin{align}
    \label{eq:entitypairs}
    \mathrm{sim}((h,t),(h^*,t^*))=\cos(\vec{h},\vec{h^*}) \cos(\vec{t},\vec{t^*})
\end{align}
Here, $\cos$ is the cosine similarity between two vectors converted to nonnegative values (i.e. $[0,1]$) using the linear transformation $(x+1)/2$.
On average, when considering 1NN, we borrow $1.3$ LDPs for each without-mention pair of entities.   
In contrast to the proposed SuperBorrow, NeighbBorrow is unsupervised and decouples entities in each pair when computing their similarity.

\section{Results}
\label{sec:Res}

\paragraph{Link Prediction:}
\autoref{tab:LinkPredict} shows the results of link prediction for different settings on FB15K237 under different KGE methods.
Two translational distance-based KGE methods (i.e. \textbf{TransE} and \textbf{RotatE}) and two semantic matching-based models (i.e. \textbf{DistMult} and \textbf{ComplEx}) are used as the KGE learning methods~\cite{rossi2021knowledge,wang2021survey}. 
We emphasize that our purpose here is \emph{not} to compare the absolute performance among those KGE methods, but to evaluate the effect of using LDPs for augmenting the KG and representing the without-mention entity-pairs via different borrowing methods.
For SuperBorrow, the optimal numbers of borrowed LDPs ($k$) determined using the validation set for TransE, DistMult, ComplEx and RotatE respectively are 30, 20, 15 and 25.

As shown in~\autoref{tab:LinkPredict}, augmenting the KG with the extracted LDPs (i.e., KG+ExtractedLDPs) significantly improves the performance for with-mention entity-pairs for all KGE methods. 
However, the performance when predicting links for without-mention entity-pairs decreases slightly for all KGE methods, except for DistMult in the KG+ExtractedLDPs setting. 
For the borrowing models, even though the co-occurrence baseline improves the prediction for without-mention set, borrowing relevant LDPs from the 1NN entity-pairs (NeighbBorrow) or the proposed supervised borrowing (SuperBorrow) reports superior results. 
We can see that the best performance for the \emph{overall} and \emph{without-mention} sets are achieved with the augmented KG using SuperBorrow, followed by NeighbBorrow. 
\begin{table}[t]
\small
        \scalebox{0.8}{
        \begin{tabular}{l c c c c c  c } \toprule
			 & \multicolumn{3}{c}{overall} & \multicolumn{3}{c}{without-mention} \\ 
			 \cmidrule(lr){2-4}  \cmidrule(lr){5-7} \\
			Model& MR  &H@3 &  H@1 & MR &  H@3 & H@1\\ \midrule
			
			\textbf{DistMult} (KG only) &4.1	&0.938 &0.856 &4.0	&0.942 &0.865\\ 
			KG+ExtractedLDPs & 2.6&	0.955&	\textbf{0.876}& 2.7&	0.957&	\textbf{0.883}\\ 
			\hdashline
			LinkAll &7.2&	0.887&	0.752&7.8&	0.880&	0.744\\
			Co-occurrence &2.4&	0.954&0.875&2.4&	0.956&	0.882\\ 
			NeighbBorrow &3.0&	0.955&	0.874&3.0&	0.956&	0.881\\ 
			SuperBorrow &\textbf{2.2}&	\textbf{0.960}&	0.875&\textbf{2.2}&	\textbf{0.962}&	0.882\\ \midrule 
			
			\textbf{ComplEx} (KG only) &3.1&	0.954&	0.900&2.8&	0.957&	0.908\\ 
			KG+ExtractedLDPs &1.9&	0.967&	0.917&1.9&	0.967&	0.922\\ 
			\hdashline
			LinkAll &4.0&	0.909&	0.812&4.3&	0.902&	0.808\\
			Co-occurrence &1.8&	0.967&	0.916&1.8&	0.967&	0.920\\ 
			NeighbBorrow &\textbf{1.7}&	\textbf{0.973}&	\textbf{0.921}&\textbf{1.7}&	\textbf{0.974}&	\textbf{0.925}\\ 
			SuperBorrow &\textbf{1.7}&	0.972&	0.917&\textbf{1.7}&	0.973&	0.922\\ 
			\bottomrule
	\end{tabular}}
    \caption{Results of relation prediction on FB15K237.}
    \label{tab:RelPredict}
\end{table}

\paragraph{Relation Prediction:} 
\autoref{tab:RelPredict} shows the accuracies for the relation prediction task.
Experimentally, the best results for this task is obtained when corrupting $r$, in addition to $h$ and $t$ corruptions, is applied to generate negative triples to train the KGE method. 
This negative sampling schedule follows the evaluation procedure of relation prediction.
As shown in the table, SuperBorrow reports the best MR and Hits@3 for DistMult KGEs, while NeibhBorrow baseline performs better than SuperBorrow with ComplEx method. 
Further results for relation prediction are in the Supplementary~\autoref{sec:appendixRelPredict}.

\paragraph{Comparisons against Prior Work:} 
We compare our proposed method against prior work, namely Feature Rich Network (\textbf{FRN})~\citep{komninos2017feature} and \textbf{Conv (E+DistMult)}~\citep{toutanova2015representing}. 
In FRN, an MLP is trained to predict the probability of a given triple being true using different types of features such as the entity types and features extracted from textual relation mentions. 
Conv(E+DistMult) represents LDPs by vectors using a convolutional neural network, and combines DistMult scoring function with that of the Entity model (E) proposed by~\citet{riedel2013relation}.
E model learns a vector for each entity and two vectors for each relation corresponding to the two arguments $\vec{r}_h$ and $\vec{r}_t$ of a relation $r$.
The scoring function of a triple in E model is defined as $\vec{h}\T\vec{r}_h+\vec{t}\T\vec{r}_t$.
The combined model (E+DistMult) is trained on a linearly weighted combination of KG triples and textual triples.
For a fair comparison, we consider the task of predicting missing tail entities and we avoid the type-constraint setting. 

As shown in~\autoref{tab:comparison}, for the overall test set of FB15K237 our models outperform both FRN and Conv models according to MRR and H@10. 
For with-mention entity-pairs, our models report higher scores compared to Conv(E+DistMult), while FRN performs best.
For with-mention entity-pairs FRN can extract rich features from the contexts of co-occurrences, which helps it to obtain superior performances.
However, both FRN and Conv models perform poorly on without-mention entity-pairs, where such contextual features are unavailable.
On the other hand, by using the proposed SuperBorrow to augment LDPs for KGs we can overcome this limitation successfully.

\begin{table}
    \centering
    \scalebox{0.7}{
    \begin{tabular}{l c c c c c c }\toprule
    & \multicolumn{2}{c}{overall} & \multicolumn{2}{c}{with-mention}&\multicolumn{2}{c}{without-mention} \\ 
			 \cmidrule(lr){2-3}  \cmidrule(lr){4-5} \cmidrule(lr){6-7} \\
			Model& MRR&H@10 & MRR& H@10 & MRR&H@10\\
			\midrule
         Conv (E+DistMult)&0.401&0.581&0.339&0.499&0.424&0.611\\
         FRN&0.403&0.620&\textbf{0.441}&\textbf{0.683}&0.387&0.595\\
         ours (DistMult)&\underline{0.460}&\textbf{0.714}&0.378&0.649&\underline{0.468}&\textbf{0.720}\\
         ours (RotatE)&\textbf{0.499}&\underline{0.712}&\underline{0.439}&\underline{0.674}&\textbf{0.504}&\underline{0.715}\\
         \bottomrule
    \end{tabular}}
    \caption{Comparisons against prior work on link prediction on FB15K237. The results for prior work are taken from the original papers. The best results are in bold, while the second best results are underlined.}
    \label{tab:comparison}
\end{table}

\section{Analysis}

\paragraph{Borrowed LDPs:} 
\begin{table*}[h]
\centering

        \scalebox{0.85}{
        \begin{tabular}{l l l} \toprule
			 Entity-pairs $(h,r,t)$&\multicolumn{2}{l}{Borrowed LDPs} \\
			 &NeighbBorrow& SuperBorrow\\
			 \midrule 
			 \emph{h}= Woodrow Wilson&\emph{h}:$\langle$-nsubj$\rangle$:joined:$\langle$dobj$\rangle$:\emph{t}&\emph{h}:$\langle$-poss$\rangle$:\emph{t}\\
			 \emph{t}= League of Nations&\emph{h}:$\langle$-nsubj$\rangle$:left:$\langle$dobj$\rangle$:\emph{t}&\emph{h}:$\langle$-nsubj$\rangle$:president:$\langle$prep$\rangle$:of:$\langle$pobj$\rangle$:\emph{t}\\
			 \emph{r}= organizations-founded&\emph{h}:$\langle$-poss$\rangle$:\emph{t}&\emph{h}:$\langle$-nsubj$\rangle$:joined:$\langle$dobj$\rangle$:\emph{t}\\
			 &&\emph{h}:$\langle$-poss$\rangle$:ambassador:$\langle$prep$\rangle$:to:$\langle$pobj$\rangle$:\emph{t}\\
			  &&\emph{h}:$\langle$-nsubj$\rangle$:member:$\langle$prep$\rangle$:of:$\langle$pobj$\rangle$:\emph{t}\\
			 \hline

			 $h$= 20th Century Fox &\emph{h}:$\langle$-nn$\rangle$:movie:$\langle$appos$\rangle$:\emph{t}&\emph{h}:$\langle$-dobj$\rangle$:released:$\langle$nsubj$\rangle$:\emph{t}\\
			 $t$= Lincoln &\emph{h}:$\langle$-nn$\rangle$:film:$\langle$nsubj$\rangle$:\emph{t}&\emph{h}:$\langle$-dobj$\rangle$:release:$\langle$nsubj$\rangle$:\emph{t}\\
			 $r$= film-distributor&\emph{h}:$\langle$-nn$\rangle$:movie:$\langle$dep$\rangle$:\emph{t}&\emph{h}:$\langle$-nsubj$\rangle$:released:$\langle$dobj$\rangle$:\emph{t}\\
			 &\emph{h}:$\langle$-poss$\rangle$:\emph{t}&\emph{h}:$\langle$-appos$\rangle$:grant:$\langle$appos$\rangle$:\emph{t}\\
			 &&\emph{h}:$\langle$-pobj$\rangle$:with:$\langle$-prep$\rangle$:\emph{t}\\
			 
			 \hline 
			 $h$= Deep Impact &\emph{h}:$\langle$-pobj$\rangle$:in:$\langle$-prep$\rangle$:work:$\langle$poss$\rangle$:\emph{t}&\emph{h}:$\langle$-dep$\rangle$:film:$\langle$poss$\rangle$:\emph{t}\\
			 $t$= Leslie Dilley&$h$:$\langle$-nn$\rangle$:fame:$\langle$-pobj$\rangle$:of:$\langle$-prep$\rangle$:$t$&\emph{h}:$\langle$vmod$\rangle$:produced:$\langle$prep$\rangle$:by:$\langle$pobj$\rangle$:\emph{t}\\
			 $r$= film-production-design-by&\emph{h}:$\langle$poss$\rangle$:\emph{t}&\emph{h}:$\langle$vmod$\rangle$:written:$\langle$prep$\rangle$:by:$\langle$pobj$\rangle$:\emph{t}\\
			 &&\emph{h}:$\langle$-dep$\rangle$:tagged:$\langle$appos$\rangle$:\emph{t}\\
			 &&\emph{h}:$\langle$-nn$\rangle$:film:$\langle$nsubj$\rangle$:\emph{t}\\
			 
			 \hline 
			 $h$= Idaho &$h$:$\langle$-nsubjpass$\rangle$:located:$\langle$prep$\rangle$:in:$\langle$pobj$\rangle$:$t$&\emph{h}:$\langle$-poss$\rangle$:\emph{t}\\
			 $t$= Christianity&\emph{h}:$\langle$-appos$\rangle$:usa:$\langle$-appos$\rangle$:\emph{t}&\emph{h}:$\langle$-amod$\rangle$:state:$\langle$prep$\rangle$:of:$\langle$pobj$\rangle$:\emph{t}\\
			 $r$= religion &\emph{h}:$\langle$-poss$\rangle$:\emph{t}&\emph{h}:$\langle$rcmod$\rangle$:plays:$\langle$dobj$\rangle$:\emph{t}\\
			 &&\emph{h}:$\langle$-dobj$\rangle$:entered:$\langle$nsubj$\rangle$:\emph{t}\\
			 &&\emph{h}:$\langle$-nn$\rangle$:date:$\langle$nn$\rangle$:\emph{t}\\

			\bottomrule
	\end{tabular}}
 
    \caption{Borrowed LDPs of selected entity-pairs. Top 5 LDPs with our borrowing method and LDPs borrowed from 3NN entity-pairs are shown. }
    \label{tab:QualitativeAnalysis}
\end{table*}
To provide examples of LDPs injected into FB15K237,~\autoref{tab:QualitativeAnalysis} shows the borrowed LDPs by NeighbBorrow and SuperBorrow for some selected entity-pairs. 
We can see that representative LDPs of various relation types are ranked at the top by SuperBorrow. 
For example, for the \emph{film-distributor} relation, NeighbBorrow selects LDPs containing specific tokens such as \emph{movie} or \emph{film}, whereas SuperBorrow retrieves LDPs that better express the target relation such as \emph{20th Century Fox}:$\langle$-dobj$\rangle$:released:$\langle$nsubj$\rangle$:\emph{Lincoln}.

\paragraph{Relation Categories:}
To better analyse the effect of the proposed SuperBorrow for KGEs, we evaluate the link prediction task on different relation categories including 1to1, 1toN, Nto1 and NtoN as defined in~\citet{bordes2013translating}.   

\autoref{tab:1toN} presents the results of predicting head entities for all KGE methods considering KG only and SuperBorrow. 
We can see that SuperBorrow achieves higher performance over the original graph on all relation categories. 
In particular, our proposal significantly boosts the performance of predicting head entities for the Nto1 relation type where all KGE methods report the lowest H@10 for the KG only setting.
Similar results are obtained for predicting the tail entities as in~\autoref{sec:appendixTailPredict}.
Overall, these results show that incorporating information from text corpora into KGs enables us to learn KGEs that encode diverse relation types. 
\begin{table}[t]
    \centering
    \scalebox{0.76}{
    \begin{tabular}{c l c c c c} \toprule
    \multicolumn{2}{l}{Method}&1to1&1toN&Nto1&NtoN\\ 
    &\# Tuples& 192& 1293& 4289& 14,696\\
    \midrule 
        \multirow{2}{*}{TransE} & KG only&0.536& 0.597&0.124& 0.418 \\
        &SuperBorrow&0.947& 0.984& 0.377&0.829\\
        \hline 
        \multirow{2}{*}{DistMult} & KG only&0.500& 0.433& 0.064& 0.371\\
        &SuperBorrow&0.922& 0.856& 0.338& 0.547\\
        \hline
        \multirow{2}{*}{ComplEx} & KG only&0.495& 0.434& 0.045& 0.368 \\
        &SuperBorrow&0.917& 0.913& 0.277& 0.601\\
        \hline
        \multirow{2}{*}{RotatE} & KG only&0.568& 0.631& 0.118& 0.388 \\
        &SuperBorrow&0.932& 0.969& 0.404& 0.722\\
        
     \bottomrule
    \end{tabular}}
    \caption{Hits@10 of tail prediction for different relation categories.}
    \label{tab:1toN}
\end{table}

\begin{figure}[t]
 \centering
 \begin{subfigure}{0.23\textwidth}
 \includegraphics[width=\textwidth]{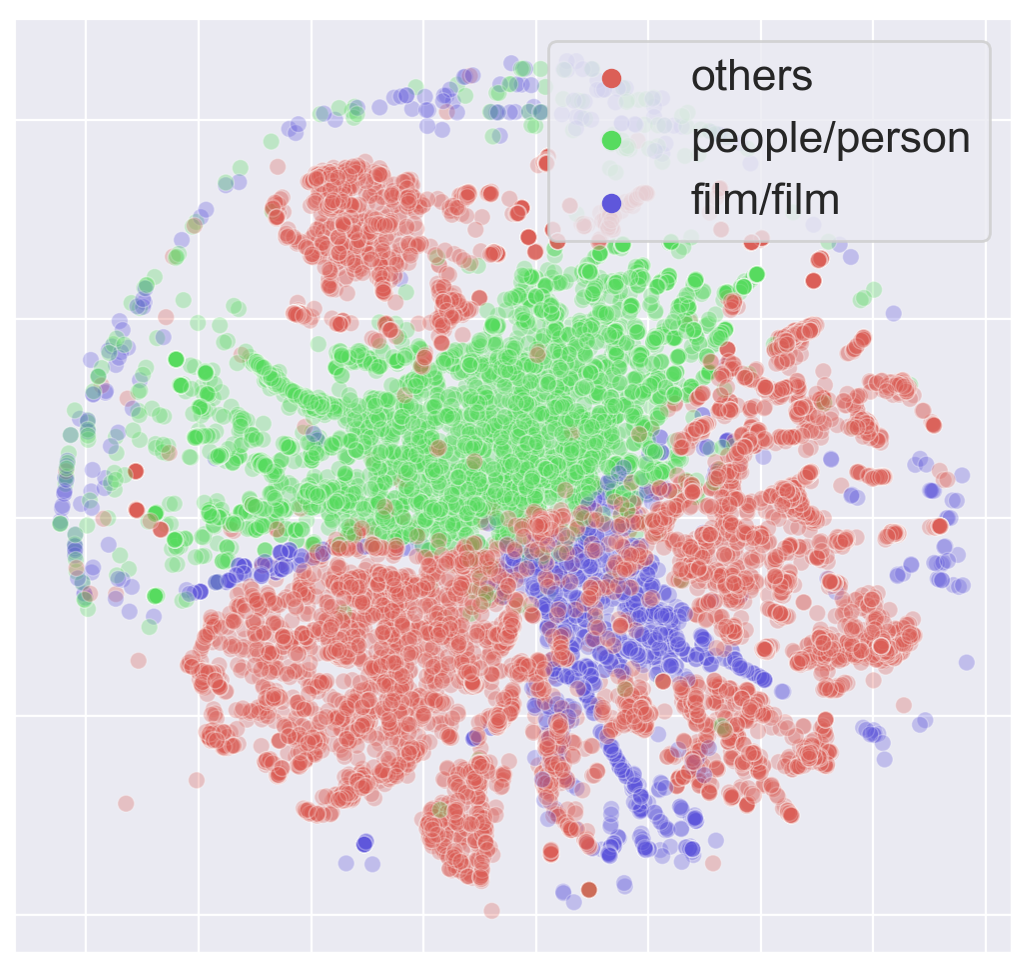}
 \caption{KG only}
 \label{fig:KGonly}
 \end{subfigure}
 \begin{subfigure}{0.23\textwidth}
 \includegraphics[width=\textwidth]{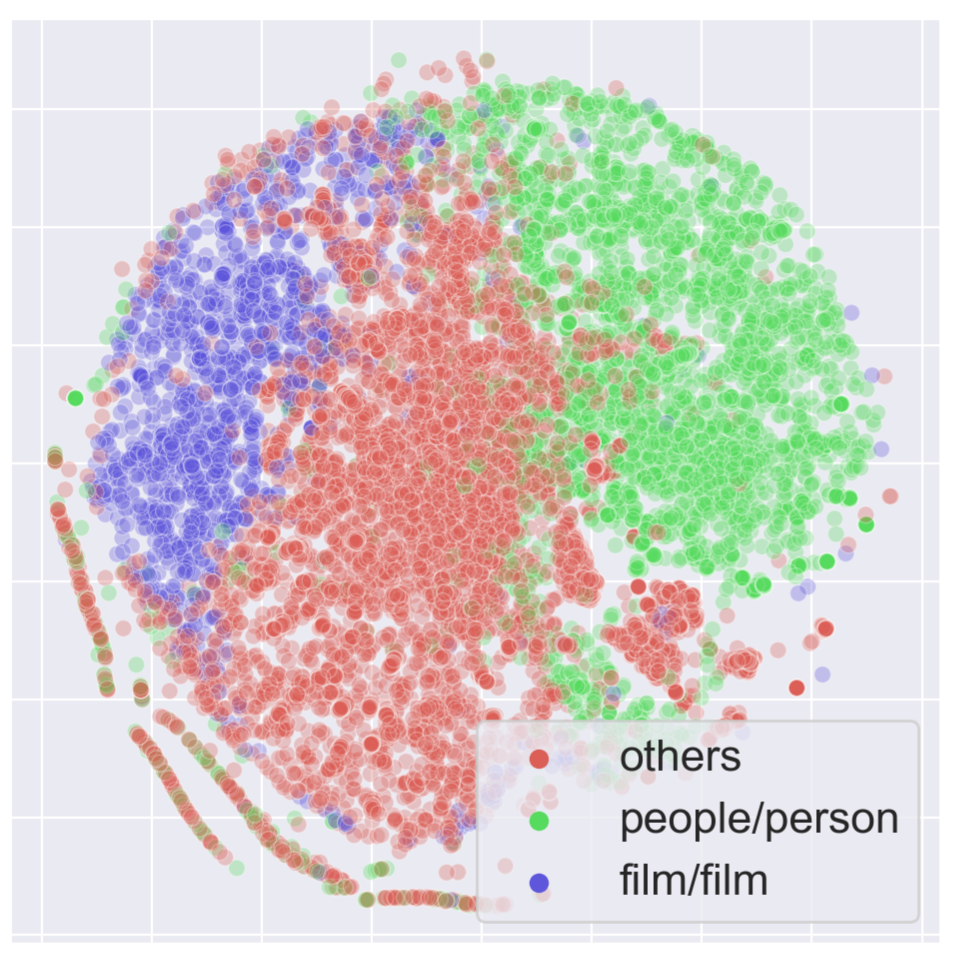}
  \caption{KG with LDPs}
 \end{subfigure}
 \caption{t-SNE plots for DistMult entity embeddings comparing (a) KG-only and (b) KG with LDPs. }
 \label{fig:tsne}
 \vspace{-4mm}
 \end{figure}

\paragraph{Visualisation of Entity Embeddings:}
In \autoref{fig:tsne}, we visualise the entity embeddings of KGonly and KG with LDPs using $t$-distributed stochastic neighbour embeddings (t-SNE)~\citep{hinton2002stochastic} method.
Relations in FB15k237 are labelled as \textrm{domain/type/property} where \textrm{domain/type} represents the type of a head entity in the relation. 
Thus, for each entity in the KG, we extract its types from all training triples where the entity acts as the head.
We label entities that belong to the two most frequent entity types, which are people/person (4,538 entities) and film/film (1,923 entities).
From \autoref{fig:tsne}, we see that the embeddings learnt from the augmented graph results in distinct clusters of the same type, compared to the clusters obtained from the KG alone.
This emphasizes the importance of using textual mentions in KGE learning. 

\section{Conclusion}
We considered the problem of representing without-mention entity-pairs in KGE learning.
Specifically, we proposed a method (SuperBorrow) to determine which LDPs to borrow from with-mention entity-pairs to augment a KG using a corpus.
Our proposed method improves the performance of several KGE learning methods in link and relation prediction tasks.

\bibliography{anthology,myrefs}

\begin{thebibliography}{}
\expandafter\ifx\csname natexlab\endcsname\relax\def\natexlab#1{#1}\fi

\bibitem[{An et~al.(2018)An, Chen, Han, and Sun}]{an2018accurate}
Bo~An, Bo~Chen, Xianpei Han, and Le~Sun. 2018.
\newblock Accurate text-enhanced knowledge graph representation learning.
\newblock In {\em Proceedings of the 2018 Conference of the North American
  Chapter of the Association for Computational Linguistics: Human Language
  Technologies, Volume 1 (Long Papers)\/}. pages 745--755.

\bibitem[{Auer et~al.(2007)Auer, Bizer, Kobilarov, Lehmann, Cyganiak, and
  Ives}]{auer2007dbpedia}
S{\"o}ren Auer, Christian Bizer, Georgi Kobilarov, Jens Lehmann, Richard
  Cyganiak, and Zachary Ives. 2007.
\newblock Dbpedia: A nucleus for a web of open data.
\newblock In {\em The semantic web\/}, Springer, pages 722--735.

\bibitem[{Bollacker et~al.(2008)Bollacker, Evans, Paritosh, Sturge, and
  Taylor}]{bollacker2008freebase}
Kurt Bollacker, Colin Evans, Praveen Paritosh, Tim Sturge, and Jamie Taylor.
  2008.
\newblock Freebase: a collaboratively created graph database for structuring
  human knowledge.
\newblock In {\em Proceedings of the 2008 ACM SIGMOD international conference
  on Management of data\/}. pages 1247--1250.

\bibitem[{Bollegala et~al.(2021)Bollegala, Hakami, Yoshida, and
  Kawarabayashi}]{bollegala-etal-2021-relwalk}
Danushka Bollegala, Huda Hakami, Yuichi Yoshida, and Ken-ichi Kawarabayashi.
  2021.
\newblock \href{https://doi.org/10.18653/v1/2021.eacl-main.133}{{R}el{W}alk - a
  latent variable model approach to knowledge graph embedding}.
\newblock In {\em Proceedings of the 16th Conference of the European Chapter of
  the Association for Computational Linguistics: Main Volume\/}. Association
  for Computational Linguistics, Online, pages 1551--1565.
\newblock
  \href{https://doi.org/10.18653/v1/2021.eacl-main.133}{https://doi.org/10.18653/v1/2021.eacl-main.133}.

\bibitem[{Bollegala et~al.(2010)Bollegala, Matsuo, and
  Ishizuka}]{Bollegala:WWW:2010}
Danushka Bollegala, Yutaka Matsuo, and Mitsuru Ishizuka. 2010.
\newblock Relational duality: Unsupervised extraction of semantic relations
  between entities on the web.
\newblock In {\em WWW 2010\/}. pages 151 -- 160.

\bibitem[{Bordes et~al.(2013)Bordes, Usunier, Garcia-Duran, Weston, and
  Yakhnenko}]{bordes2013translating}
Antoine Bordes, Nicolas Usunier, Alberto Garcia-Duran, Jason Weston, and Oksana
  Yakhnenko. 2013.
\newblock Translating embeddings for modeling multi-relational data.
\newblock In {\em Neural Information Processing Systems (NIPS)\/}. pages 1--9.

\bibitem[{Chang et~al.(2014)Chang, Yih, Yang, and Meek}]{chang2014typed}
Kai-Wei Chang, Wen-tau Yih, Bishan Yang, and Christopher Meek. 2014.
\newblock Typed tensor decomposition of knowledge bases for relation
  extraction.
\newblock In {\em Proceedings of the 2014 Conference on Empirical Methods in
  Natural Language Processing (EMNLP)\/}. pages 1568--1579.

\bibitem[{Chen and Manning(2014)}]{chen2014fast}
Danqi Chen and Christopher~D Manning. 2014.
\newblock A fast and accurate dependency parser using neural networks.
\newblock In {\em Proceedings of the 2014 conference on empirical methods in
  natural language processing (EMNLP)\/}. pages 740--750.

\bibitem[{Das et~al.(2017)Das, Zaheer, Reddy, and McCallum}]{das2017question}
Rajarshi Das, Manzil Zaheer, Siva Reddy, and Andrew McCallum. 2017.
\newblock Question answering on knowledge bases and text using universal schema
  and memory networks.
\newblock {\em arXiv preprint arXiv:1704.08384\/} .

\bibitem[{Duchi et~al.(2011)Duchi, Hazan, and Singer}]{duchi2011adaptive}
John Duchi, Elad Hazan, and Yoram Singer. 2011.
\newblock Adaptive subgradient methods for online learning and stochastic
  optimization.
\newblock {\em Journal of machine learning research\/} 12(7).

\bibitem[{Ehrlinger and W{\"o}{\ss}(2016)}]{ehrlinger2016towards}
Lisa Ehrlinger and Wolfram W{\"o}{\ss}. 2016.
\newblock Towards a definition of knowledge graphs.
\newblock {\em SEMANTiCS (Posters, Demos, SuCCESS)\/} 48(1-4):2.

\bibitem[{Gabrilovich et~al.(2013)Gabrilovich, Ringgaard, and
  Subramanya}]{gabrilovich2013facc1}
Evgeniy Gabrilovich, Michael Ringgaard, and Amarnag Subramanya. 2013.
\newblock Facc1: Freebase annotation of clueweb corpora.

\bibitem[{Hakami and Bollegala(2019)}]{hakami2019context}
Huda Hakami and Danushka Bollegala. 2019.
\newblock Context-guided self-supervised relation embeddings.
\newblock In {\em International Conference of the Pacific Association for
  Computational Linguistics\/}. Springer, pages 67--78.

\bibitem[{Han et~al.(2018)Han, Cao, Xin, Lin, Liu, Sun, and Li}]{han2018openke}
Xu~Han, Shulin Cao, Lv~Xin, Yankai Lin, Zhiyuan Liu, Maosong Sun, and Juanzi
  Li. 2018.
\newblock Openke: An open toolkit for knowledge embedding.
\newblock In {\em Proceedings of EMNLP\/}.

\bibitem[{Hinton and Roweis(2002)}]{hinton2002stochastic}
Geoffrey Hinton and Sam~T Roweis. 2002.
\newblock Stochastic neighbor embedding.
\newblock In {\em NIPS\/}. Citeseer, volume~15, pages 833--840.

\bibitem[{Joshi et~al.(2018)Joshi, Choi, Levy, Weld, and
  Zettlemoyer}]{joshi2018pair2vec}
Mandar Joshi, Eunsol Choi, Omer Levy, Daniel~S Weld, and Luke Zettlemoyer.
  2018.
\newblock pair2vec: Compositional word-pair embeddings for cross-sentence
  inference.
\newblock {\em arXiv preprint arXiv:1810.08854\/} .

\bibitem[{Komninos and Manandhar(2017)}]{komninos2017feature}
Alexandros Komninos and Suresh Manandhar. 2017.
\newblock Feature-rich networks for knowledge base completion.
\newblock In {\em Proceedings of the 55th Annual Meeting of the Association for
  Computational Linguistics (Volume 2: Short Papers)\/}. pages 324--329.

\bibitem[{Kroetsch and Weikum(2016)}]{kroetsch2016special}
M~Kroetsch and G~Weikum. 2016.
\newblock Special issue on knowledge graphs.
\newblock {\em Journal of Web Semantics\/} 37(38):53--54.

\bibitem[{Long et~al.(2016)Long, Lowe, Cheung, and Precup}]{long2016leveraging}
Teng Long, Ryan Lowe, Jackie Chi~Kit Cheung, and Doina Precup. 2016.
\newblock Leveraging lexical resources for learning entity embeddings in
  multi-relational data.
\newblock {\em arXiv preprint arXiv:1605.05416\/} .

\bibitem[{Lu et~al.(2020)Lu, Cong, and Huang}]{lu2020utilizing}
Fengyuan Lu, Peijin Cong, and Xinli Huang. 2020.
\newblock Utilizing textual information in knowledge graph embedding: A survey
  of methods and applications.
\newblock {\em IEEE Access\/} 8:92072--92088.

\bibitem[{Min et~al.(2013)Min, Grishman, Wan, Wang, and
  Gondek}]{min2013distant}
Bonan Min, Ralph Grishman, Li~Wan, Chang Wang, and David Gondek. 2013.
\newblock Distant supervision for relation extraction with an incomplete
  knowledge base.
\newblock In {\em Proceedings of the 2013 Conference of the North American
  Chapter of the Association for Computational Linguistics: Human Language
  Technologies\/}. pages 777--782.

\bibitem[{Minervini et~al.(2015)Minervini, d'Amato, Fanizzi, and
  Esposito}]{minervini2015efficient}
Pasquale Minervini, Claudia d'Amato, Nicola Fanizzi, and Floriana Esposito.
  2015.
\newblock Efficient learning of entity and predicate embeddings for link
  prediction in knowledge graphs.
\newblock In {\em URSW@ ISWC\/}. pages 26--37.

\bibitem[{Nickel et~al.(2015)Nickel, Murphy, Tresp, and
  Gabrilovich}]{nickel2015review}
Maximilian Nickel, Kevin Murphy, Volker Tresp, and Evgeniy Gabrilovich. 2015.
\newblock A review of relational machine learning for knowledge graphs.
\newblock {\em Proceedings of the IEEE\/} 104(1):11--33.

\bibitem[{Nickel et~al.(2011)Nickel, Tresp, and Kriegel}]{nickel2011three}
Maximilian Nickel, Volker Tresp, and Hans-Peter Kriegel. 2011.
\newblock A three-way model for collective learning on multi-relational data.
\newblock In {\em Icml\/}.

\bibitem[{Paulheim(2017)}]{paulheim2017knowledge}
Heiko Paulheim. 2017.
\newblock Knowledge graph refinement: A survey of approaches and evaluation
  methods.
\newblock {\em Semantic web\/} 8(3):489--508.

\bibitem[{Reimers and Gurevych(2019)}]{reimers2019sentence}
Nils Reimers and Iryna Gurevych. 2019.
\newblock Sentence-bert: Sentence embeddings using siamese bert-networks.
\newblock {\em arXiv preprint arXiv:1908.10084\/} .

\bibitem[{Rezayi et~al.(2021)Rezayi, Zhao, Kim, Rossi, Lipka, and
  Li}]{rezayi-etal-2021-edge}
Saed Rezayi, Handong Zhao, Sungchul Kim, Ryan Rossi, Nedim Lipka, and Sheng Li.
  2021.
\newblock \href{https://doi.org/10.18653/v1/2021.naacl-main.221}{Edge:
  Enriching knowledge graph embeddings with external text}.
\newblock In {\em Proceedings of the 2021 Conference of the North American
  Chapter of the Association for Computational Linguistics: Human Language
  Technologies\/}. Association for Computational Linguistics, Online, pages
  2767--2776.
\newblock
  \href{https://doi.org/10.18653/v1/2021.naacl-main.221}{https://doi.org/10.18653/v1/2021.naacl-main.221}.

\bibitem[{Riedel et~al.(2013)Riedel, Yao, McCallum, and
  Marlin}]{riedel2013relation}
Sebastian Riedel, Limin Yao, Andrew McCallum, and Benjamin~M Marlin. 2013.
\newblock Relation extraction with matrix factorization and universal schemas.
\newblock In {\em Proceedings of the 2013 Conference of the North American
  Chapter of the Association for Computational Linguistics: Human Language
  Technologies\/}. pages 74--84.

\bibitem[{Rossi et~al.(2021)Rossi, Barbosa, Firmani, Matinata, and
  Merialdo}]{rossi2021knowledge}
Andrea Rossi, Denilson Barbosa, Donatella Firmani, Antonio Matinata, and Paolo
  Merialdo. 2021.
\newblock Knowledge graph embedding for link prediction: A comparative
  analysis.
\newblock {\em ACM Transactions on Knowledge Discovery from Data (TKDD)\/}
  15(2):1--49.

\bibitem[{Rosso et~al.(2019)Rosso, Yang, and
  Cudre-Mauroux}]{rosso2019revisiting}
Paolo Rosso, Dingqi Yang, and Philippe Cudre-Mauroux. 2019.
\newblock Revisiting text and knowledge graph joint embeddings: The amount of
  shared information matters!
\newblock In {\em 2019 IEEE International Conference on Big Data (Big Data)\/}.
  IEEE, pages 2465--2473.

\bibitem[{Shen et~al.(2014)Shen, Wang, and Han}]{shen2014entity}
Wei Shen, Jianyong Wang, and Jiawei Han. 2014.
\newblock Entity linking with a knowledge base: Issues, techniques, and
  solutions.
\newblock {\em IEEE Transactions on Knowledge and Data Engineering\/}
  27(2):443--460.

\bibitem[{Socher et~al.(2013)Socher, Chen, Manning, and
  Ng}]{socher2013reasoning}
Richard Socher, Danqi Chen, Christopher~D Manning, and Andrew Ng. 2013.
\newblock Reasoning with neural tensor networks for knowledge base completion.
\newblock In {\em Advances in neural information processing systems\/}. pages
  926--934.

\bibitem[{Suchanek et~al.(2007)Suchanek, Kasneci, and
  Weikum}]{suchanek2007yago}
Fabian~M Suchanek, Gjergji Kasneci, and Gerhard Weikum. 2007.
\newblock Yago: a core of semantic knowledge.
\newblock In {\em Proceedings of the 16th international conference on World
  Wide Web\/}. pages 697--706.

\bibitem[{Sun et~al.(2019)Sun, Deng, Nie, and Tang}]{sun2019rotate}
Zhiqing Sun, Zhi-Hong Deng, Jian-Yun Nie, and Jian Tang. 2019.
\newblock Rotate: Knowledge graph embedding by relational rotation in complex
  space.
\newblock {\em arXiv preprint arXiv:1902.10197\/} .

\bibitem[{Sydorova et~al.(2019)Sydorova, Poerner, and
  Roth}]{sydorova2019interpretable}
Alona Sydorova, Nina Poerner, and Benjamin Roth. 2019.
\newblock Interpretable question answering on knowledge bases and text.
\newblock {\em arXiv preprint arXiv:1906.10924\/} .

\bibitem[{Tang et~al.(2019)Tang, Chen, Cui, and Wei}]{tang2019knowledge}
Xing Tang, Ling Chen, Jun Cui, and Baogang Wei. 2019.
\newblock Knowledge representation learning with entity descriptions,
  hierarchical types, and textual relations.
\newblock {\em Information Processing \& Management\/} 56(3):809--822.

\bibitem[{Toutanova and Chen(2015)}]{toutanova2015observed}
Kristina Toutanova and Danqi Chen. 2015.
\newblock Observed versus latent features for knowledge base and text
  inference.
\newblock In {\em Proceedings of the 3rd workshop on continuous vector space
  models and their compositionality\/}. pages 57--66.

\bibitem[{Toutanova et~al.(2015)Toutanova, Chen, Pantel, Poon, Choudhury, and
  Gamon}]{toutanova2015representing}
Kristina Toutanova, Danqi Chen, Patrick Pantel, Hoifung Poon, Pallavi
  Choudhury, and Michael Gamon. 2015.
\newblock Representing text for joint embedding of text and knowledge bases.
\newblock In {\em Proceedings of the 2015 conference on empirical methods in
  natural language processing\/}. pages 1499--1509.

\bibitem[{Trouillon et~al.(2016)Trouillon, Welbl, Riedel, Gaussier, and
  Bouchard}]{trouillon2016complex}
Th{\'e}o Trouillon, Johannes Welbl, Sebastian Riedel, {\'E}ric Gaussier, and
  Guillaume Bouchard. 2016.
\newblock Complex embeddings for simple link prediction.
\newblock In {\em International Conference on Machine Learning\/}. PMLR, pages
  2071--2080.

\bibitem[{Veira et~al.(2019)Veira, Keng, Padmanabhan, and
  Veneris}]{veira2019unsupervised}
Neil Veira, Brian Keng, Kanchana Padmanabhan, and Andreas~G Veneris. 2019.
\newblock Unsupervised embedding enhancements of knowledge graphs using textual
  associations.
\newblock In {\em IJCAI\/}. pages 5218--5225.

\bibitem[{Wang et~al.(2021)Wang, Qiu, and Wang}]{wang2021survey}
Meihong Wang, Linling Qiu, and Xiaoli Wang. 2021.
\newblock A survey on knowledge graph embeddings for link prediction.
\newblock {\em Symmetry\/} 13(3):485.

\bibitem[{Wang et~al.(2017)Wang, Mao, Wang, and Guo}]{wang2017knowledge}
Quan Wang, Zhendong Mao, Bin Wang, and Li~Guo. 2017.
\newblock Knowledge graph embedding: A survey of approaches and applications.
\newblock {\em IEEE Transactions on Knowledge and Data Engineering\/}
  29(12):2724--2743.

\bibitem[{Wang et~al.(2019{\natexlab{a}})Wang, Jiang
  et~al.}]{wang2019knowledge}
Shirui Wang, Chao Jiang, et~al. 2019{\natexlab{a}}.
\newblock Knowledge graph embedding with interactive guidance from entity
  descriptions.
\newblock {\em IEEE Access\/} 7:156686--156693.

\bibitem[{Wang et~al.(2019{\natexlab{b}})Wang, Zhang, and Xie}]{wang2019model}
Yashen Wang, Huanhuan Zhang, and Haiyong Xie. 2019{\natexlab{b}}.
\newblock A model of text-enhanced knowledge graph representation learning with
  collaborative attention.
\newblock In {\em Asian Conference on Machine Learning\/}. PMLR, pages
  220--235.

\bibitem[{Wang et~al.(2014)Wang, Zhang, Feng, and Chen}]{wang2014knowledge}
Zhen Wang, Jianwen Zhang, Jianlin Feng, and Zheng Chen. 2014.
\newblock Knowledge graph and text jointly embedding.
\newblock In {\em Proceedings of the 2014 conference on empirical methods in
  natural language processing (EMNLP)\/}. pages 1591--1601.

\bibitem[{Wang et~al.(2016)Wang, Li, Liu, and Tang}]{wang2016text}
Zhigang Wang, Juanzi Li, Zhiyuan Liu, and Jie Tang. 2016.
\newblock Text-enhanced representation learning for knowledge graph.
\newblock In {\em Proceedings of International Joint Conference on Artificial
  Intelligent (IJCAI)\/}. pages 4--17.

\bibitem[{Warstadt et~al.(2020)Warstadt, Zhang, Li, Liu, and
  Bowman}]{warstadt2020learning}
Alex Warstadt, Yian Zhang, Haau-Sing Li, Haokun Liu, and Samuel~R Bowman. 2020.
\newblock Learning which features matter: Roberta acquires a preference for
  linguistic generalizations (eventually).
\newblock {\em arXiv preprint arXiv:2010.05358\/} .

\bibitem[{Washio and Kato(2018)}]{washio2018neural}
Koki Washio and Tsuneaki Kato. 2018.
\newblock Neural latent relational analysis to capture lexical semantic
  relations in a vector space.
\newblock {\em arXiv preprint arXiv:1809.03401\/} .

\bibitem[{Weston et~al.(2013)Weston, Bordes, Yakhnenko, and
  Usunier}]{weston2013connecting}
Jason Weston, Antoine Bordes, Oksana Yakhnenko, and Nicolas Usunier. 2013.
\newblock Connecting language and knowledge bases with embedding models for
  relation extraction.
\newblock {\em arXiv preprint arXiv:1307.7973\/} .

\bibitem[{Xie et~al.(2016)Xie, Liu, Jia, Luan, and Sun}]{xie2016representation}
Ruobing Xie, Zhiyuan Liu, Jia Jia, Huanbo Luan, and Maosong Sun. 2016.
\newblock Representation learning of knowledge graphs with entity descriptions.
\newblock In {\em Proceedings of the AAAI Conference on Artificial
  Intelligence\/}. volume~30.

\bibitem[{Xu et~al.(2020)Xu, Bao, and Wang}]{xu2020knowledge}
Hongcai Xu, Junpeng Bao, and Junqing Wang. 2020.
\newblock Knowledge-graph based proactive dialogue generation with improved
  meta-learning.
\newblock In {\em 2020 2nd International Conference on Image Processing and
  Machine Vision\/}. pages 40--46.

\bibitem[{Xu et~al.(2016)Xu, Chen, Qiu, and Huang}]{xu2016knowledge}
Jiacheng Xu, Kan Chen, Xipeng Qiu, and Xuanjing Huang. 2016.
\newblock Knowledge graph representation with jointly structural and textual
  encoding.
\newblock {\em arXiv preprint arXiv:1611.08661\/} .

\bibitem[{Yang et~al.(2014)Yang, Yih, He, Gao, and Deng}]{yang2014embedding}
Bishan Yang, Wen-tau Yih, Xiaodong He, Jianfeng Gao, and Li~Deng. 2014.
\newblock Embedding entities and relations for learning and inference in
  knowledge bases.
\newblock {\em arXiv preprint arXiv:1412.6575\/} .

\bibitem[{Zhong et~al.(2015)Zhong, Zhang, Wang, Wan, and
  Chen}]{zhong2015aligning}
Huaping Zhong, Jianwen Zhang, Zhen Wang, Hai Wan, and Zheng Chen. 2015.
\newblock Aligning knowledge and text embeddings by entity descriptions.
\newblock In {\em Proceedings of the 2015 Conference on Empirical Methods in
  Natural Language Processing\/}. pages 267--272.

\end{thebibliography}
\bibliographystyle{acl_natbib}

\appendix
\section*{Appendix}
\section{Relation Prediction}
\label{sec:appendixRelPredict}
Relation prediction results for all the KGE methods are shown in~\autoref{tab:RelPrediAll}. 
As we see, unlike semantic matching-based KGE models, incorporating LDPs into the KG do not improve relation prediction for translational distance-based KGE methods (TransE and RotatE). 
For KG+ExtractedLDPs embeddings, the performance for with-mention set decreases by $0.045$ and $0.012$ on average for MRR and H@\{10,3,1\}, for TransE and RotatE respectively. 
In-depth analysis for this observation can be conducted in future research. 
\begin{table*}[h]
\small

        \scalebox{0.82}{
        \begin{tabular}{l c c c c c  c c c c c c c c c c} \toprule
			 & \multicolumn{5}{c}{overall} & \multicolumn{5}{c}{with-mention}&\multicolumn{5}{c}{without-mention} \\ 
			 \cmidrule(lr){2-6}  \cmidrule(lr){7-11} \cmidrule(lr){12-16} \\
			Model& MRR& MR & H@10&H@3&H@1 & MRR& MR & H@10&H@3&H@1 & MRR& MR & H@10&H@3&H@1\\ \midrule
			\textbf{TransE} (KG only) &0.961&	1.6&	0.992&	0.980&	0.940&	0.919&	1.9&	0.988&	0.958&	0.875&	0.967&	1.5&	0.993&	0.983&	0.949\\ 
			KG+ExtractedLDPs &0.934&	1.6&	0.994&	0.967&	0.899&	0.860&	1.7&	0.991&	0.926&	0.789&	0.944&	1.5&	0.994&	0.973&	0.914 \\
			\hdashline
			LinkAll &0.932&	1.5&	0.993&	0.955&	0.899&	0.845&	1.9&	0.985&	0.887&	0.778&	0.944&	1.4&	0.994&	0.964&	0.916\\
			Co-occurrence&0.925&	1.6&	0.993&	0.962&	0.887&	0.863&	1.8&	0.990&	0.931&	0.793&	0.933&	1.5&	0.993&	0.967&	0.899 \\
			NeighbBorrow&0.927&	1.5&	0.994&	0.964&	0.888&	0.862&	1.7&	0.993&	0.929&	0.791&	0.936&	1.5&	0.994&	0.969&	0.901\\
			SuperBorrow&0.925&	1.5&	0.993&	0.963&	0.886&	0.868&	1.8&	0.990&	0.926&	0.802&	0.933&	1.5&	0.994&	0.968&	0.897	\\
			\midrule
			\\
			\textbf{DistMult} (KG only)&0.901&	4.1&	0.968&	0.938&	0.856&	0.855&	4.5&	0.959&	0.914&	0.789&	0.907&	4.0&	0.969&	0.942&	0.865 \\
			KG+ExtractedLDPs&0.918&	2.6&	0.980&	0.955&	0.876&	0.887&	2.4&	0.982&	0.940&	0.826&	0.922&	2.7&	0.980&	0.957&	0.883\\
			\hdashline
			LinkAll &0.825&	7.2&	0.940&	0.887&	0.752&	0.883&	2.1&	0.986&	0.944&	0.813&	0.818&	7.8&	0.934&	0.880&	0.744 \\
			Co-occurrence&0.918&	2.4&	0.979&	0.954&	0.875&	0.89&	2.0&	0.980&	0.942&	0.831&	0.921&	2.4&	0.979&	0.956&	0.882 \\
			NeighbBorrow&0.917&	3.0&	0.979&	0.955&	0.874&	0.883&	2.7&	0.976&	0.942&	0.819&	0.921&	3.0&	0.979&	0.956&	0.881\\
			SuperBorrow&0.920&	2.2&	0.984&	0.960&	0.875&	0.885&	2.2&	0.980&	0.943&	0.822&	0.924&	2.2&	0.985&	0.962&	0.882 \\
			\midrule
			\\
			\textbf{ComplEx} (KG only)&0.929&	3.1&	0.977&	0.954&	0.900&	0.884&	4.8&	0.962&	0.925&	0.835&	0.935&	2.8&	0.980&	0.957&	0.908\\
			KG+ExtractedLDPs&0.944&	1.9&	0.987&	0.967&	0.917&	0.921&	1.7&	0.986&	0.962&	0.877&	0.947&	1.9&	0.987&	0.967&	0.922\\
			\hdashline
			LinkAll & 0.867&	4.0&	0.955&	0.909&	0.812&	0.906&	1.8&	0.988&	0.960&	0.848&	0.861&	4.3&	0.951&	0.902&	0.808\\
			Co-occurrence&0.944&	1.8&	0.987&	0.967&	0.916&	0.930&	1.7&	0.989&	0.965&	0.892&	0.946&	1.8&	0.987&	0.967&	0.920 \\
			NeighbBorrow&0.948&	1.7&	0.989&	0.973&	0.921&	0.925&	1.9&	0.987&	0.965&	0.884&	0.951&	1.7&	0.989&	0.974&	0.925\\
			SuperBorrow&0.946&	1.7&	0.990&	0.972&	0.917&	0.922&	1.8&	0.987&	0.962&	0.879&	0.949&	1.7&	0.990&	0.973&	0.922\\
		    \midrule
		    \\
			\textbf{RotatE} (KG only) &0.972&	1.4&	0.996&	0.990&	0.954&	0.945&	1.3&	0.993&	0.981&	0.910&	0.976&	1.4&	0.997&	0.991&	0.960\\
			KG+ExtractedLDPs&0.967&	1.3&	0.995&	0.988&	0.945&	0.933&	1.5&	0.983&	0.974&	0.892&	0.971&	1.2&	0.996&	0.990&	0.952\\
			\hdashline
			LinkAll &0.958&	1.3&	0.995&	0.984&	0.931&	0.923&	1.6&	0.983&	0.964&	0.879&	0.963&	1.3&	0.996&	0.987&	0.938 \\
			Co-occurrence&0.964&	1.3&	0.994&	0.985&	0.943&	0.931&	1.4&	0.987&	0.970&	0.892&	0.969&	1.3&	0.995&	0.987&	0.949\\
			NeighbBorrow&0.964&	1.3&	0.995&	0.985&	0.941&	0.933&	1.5&	0.985&	0.971&	0.894&	0.968&	1.2&	0.996&	0.987&	0.948\\
			SuperBorrow&0.964&	1.2&	0.995&	0.986&	0.941&	0.931&	1.5&	0.985&	0.972&	0.892&	0.968&	1.2&	0.996&	0.988&	0.947\\

			\bottomrule
	\end{tabular}}
	\caption{Relation predictino on FB15K237. }
    \label{tab:RelPrediAll}
\end{table*}

\section{Tail Prediction for Relation Categories}
\label{sec:appendixTailPredict}
\autoref{tab:1toNtailpredict} presents Hits@10 for tail prediction considering 1to1, 1toN, Nto1 and NtoN relation categories. 
As we see, SuperBorrow embeddings obtain the best results for all KGE methods and all the relation categories. 
\begin{table}[H]
    \centering
    \scalebox{0.75}{
    \begin{tabular}{c l c c c c} \toprule
    \multicolumn{2}{l}{Method}&1to1&1toN&Nto1&NtoN\\ 
    &\# Tuples& 192& 1293& 4289& 14,696\\
    \midrule 
        \multirow{2}{*}{TransE} & KG only&0.547& 0.097& 0.851& 0.574  \\
        &SuperBorrow&0.943& 0.647& 0.980& 0.907\\
        \hline 
        \multirow{2}{*}{DistMult} & KG only&0.521& 0.055& 0.774& 0.507\\
        &SuperBorrow&0.880& 0.424& 0.898& 0.657\\
        \hline
        \multirow{2}{*}{ComplEx}& KG only & 0.500& 0.034& 0.787& 0.518\\
        &SuperBorrow&0.869& 0.456& 0.964& 0.753\\
        \hline
        \multirow{2}{*}{RotatE} &KG only&0.536& 0.107& 0.855& 0.561  \\
        &SuperBorrow&0.927& 0.731& 0.983& 0.853\\
        
     \bottomrule
    \end{tabular}}
    \caption{Hits@10 of tail prediction for different relation categories.}
    \label{tab:1toNtailpredict}
\end{table}
\section{Training KGE Methods}
For reproducability, we list the hyperparameter setting to train KGE methods in~\autoref{tab:hyperparameter}.
AdaGrad~\citep{duchi2011adaptive} with 100 batches is used to learn KGEs.
\autoref{tab:trainTime} shows the training time (in hours) to train KGE methods for KG only and SuperBorrow using OpenKE-Pytorch tool~\citep{han2018openke}. 

\begin{table*}[h]
    \centering
    \scalebox{0.85}{
    \begin{tabular}{c|c c c c c c }
    \toprule
         KGE Method&learning rate&embedding dimension& negative samples&loss function&margin& epochs  \\
         \hline
         TransE&1.0&300&25&Margin loss&5.0&1000 \\
         DistMult&0.5&300&25&SoftPlus loss&-&1000\\
         ComplEx&0.5&100&25&SoftPlus loss&-&1000\\
         RotatE&2e-5&300&25&SigmoidLoss&9.0&1000\\
        \bottomrule
    \end{tabular}}
    \caption{The hyperparameter setting for KGE methods on link prediction task.}
    \label{tab:hyperparameter}
\end{table*}

\begin{table}[H]
    \centering
    \scalebox{0.85}{
    \begin{tabular}{c c c c}
    \toprule
         \multicolumn{2}{c}{Method}&\#Train tuples &Time (h)  \\
         \hline
         \multirow{2}{*}{TransE} & KG only&272,115& 0.42 \\
         &SuperBorrow&1,217,294&1.58\\
         \hline
         \multirow{2}{*}{DistMult} & KG only&272,115&0.78 \\
         &SuperBorrow&1,036,904&2.67\\
         \hline
         \multirow{2}{*}{ComplEx} & KG only&272,115& 0.69\\
         &SuperBorrow&946,709&2.11\\
         \hline
         \multirow{2}{*}{RotatE} & KG only&272,115&1.11 \\
         &SuperBorrow&1,127,099&4.13\\
         \bottomrule
    \end{tabular}}
    \caption{Training time on FB15K237 in hours.}
    \label{tab:trainTime}
    
\end{table}
\vfill
\end{document}